\documentclass[conference]{IEEEtran}  
\usepackage{graphicx}
\IEEEoverridecommandlockouts                              

\makeatletter
\newcommand{\linebreakand}{%
  \end{@IEEEauthorhalign}
  \hfill\mbox{}\par
  \mbox{}\hfill\begin{@IEEEauthorhalign}
}

\makeatother
\title{\LARGE \bf
Comparative Analysis of Multiple Deep CNN Models for Waste Classification
}

\author{
    \IEEEauthorblockN{Dipesh Gyawali\textsuperscript1,    Alok Regmi\textsuperscript2,  Aatish Shakya\textsuperscript3,  Ashish Gautam\textsuperscript4,  Surendra Shrestha\textsuperscript5}
    \IEEEauthorblockA{\textit{Department of Electronics and Computer Engineering, Pulchowk Campus}\\
    Institute of Engineering, Tribhuvan University\\
    Lalitpur, Nepal
     \\}
     \IEEEauthorblockN{E-mail: (\textsuperscript1072bex418, \textsuperscript2072bex403,
     \textsuperscript3072bex401,
     \textsuperscript4072bex406,
     \textsuperscript5surendra)@ioe.edu.np
    \\}}

\begin{document}

\maketitle
\thispagestyle{empty}
\pagestyle{empty}

\begin{abstract}

Waste is a wealth in a wrong place. Our research focuses on analyzing possibilities for automatic waste sorting and collecting in such a way that helps it for further recycling process. Various approaches are being practiced managing waste but not efficient and require human intervention. The automatic waste segregation would fit in to fill the gap. The project tested well known Deep Learning Network architectures for waste classification with dataset combined from own endeavors and Trash Net. The convolutional neural network is used for image classification. The hardware built in the form of dustbin is used to segregate those wastes into different compartments. Without the human exercise in segregating those waste products, the study would save the precious time and would introduce the automation in the area of waste management. Municipal solid waste is a huge, renewable source of energy. The situation is win-win for both government, society and industrialists. Because of fine-tuning of the ResNet18 Network, the best validation accuracy was found to be 87.8\%.

\end{abstract}


\section{Introduction}

 With the tremendous increase in population worldwide, the solid waste production has increased tremendously. Improper waste management has an adverse effect in economic, public health and the environment. The proper waste management of solid waste for any growing cities, towns has been a headache all around the world. Effective waste recycling is both economic and environmentally beneficial. It can help in recovering raw resource, preserving energy, mitigating greenhouse gaseous emission, water pollution, reducing new landfills, etc. 
\par

In developing country, Municipal Solid Waste recycling relies on household separation via scavengers and collectors who trade the recyclables for profits. In developed countries, communities are more involved in recycling program. Several techniques, such as mechanical sorting and chemical sorting, are available in developed countries for automatic waste sorting.
\par

Deep learning is a class of machine learning algorithm that uses multiple layer of data representation and feature extraction. Using convolution neural network[1], a class of deep feed-forward artificial neural network has been applied successively to analyze the image. Thus, in waste segregation using deep learning involves acquiring images from camera with detection[2], object recognition, prediction and classification[3] into categories as biodegradable and non-biodegradable.

 \begin{figure}
  \centering
  \includegraphics[scale=0.15]{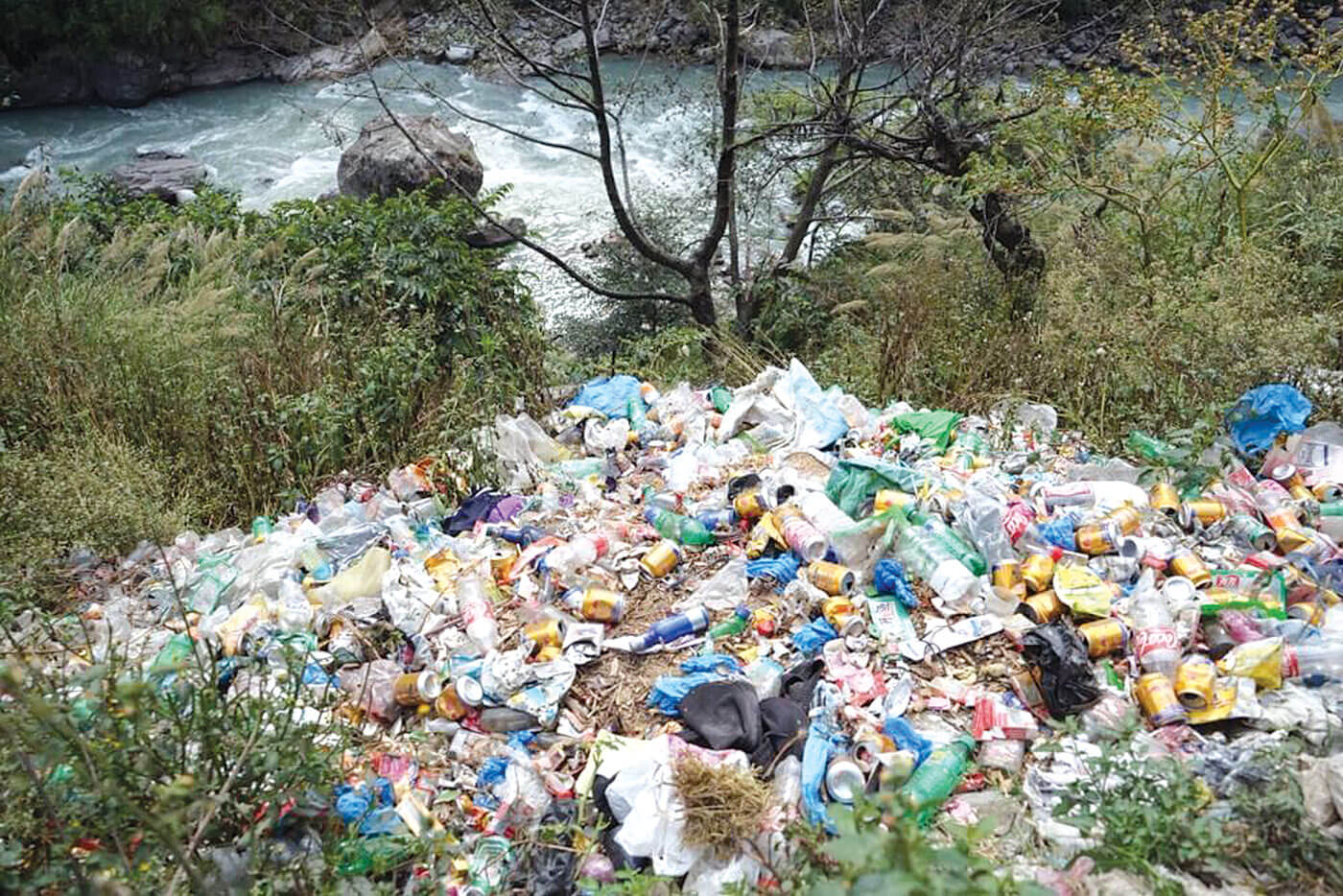}
  \caption{Different types of waste residing on the land}
\end{figure}

\medskip

\section{{Related Works}}
There has been many theoretical projects and several experimental projects individually done by garbage companies as well. From the researches that have been done, the dataset is usually the problematic part in most of the cases. The need of a dataset that properly encompasses the problem set is of extreme importance. The recent developments in convolutional neural networks (CNN) has created an awesome environment for learning the features from images resulting in image classification, segmentation and detection [1]. Therefore, such way seems to be the best way to classify wastes on a whole. Awe et al. [2] propose an experimental project using a Faster R-CNN model to classify waste into three categories: paper, recycling, and landﬁll that achieved a mean average precision of 68\%. Thung and Yang [3] deployed support vector machine (SVM) and a convolutional neural network (CNN) to classify waste into six categories. It achieves an accuracy rate of 63\% for SVM and 73\% for CNN. Rad et al. [4] developed a GoogLeNet-based vision application to localize and classify urban wastes. The study claims to have an accuracy rate ranging from 63\% to 77\% for diﬀerent waste types. Donovan [5] proposed to use Google’s TensorFlow and camera capturing to automatically sort waste objects as compost and recyclable. However, as a conceptual project, there is no experimental result so far. Mittal et al. [6] designed a project to detect whether an image contains garbage or not where he employs the pretrained AlexNet model and achieves a mean accuracy of 87.69\%. Mittal’s project aims at segmenting garbage in an image without providing functions of waste classiﬁcation. An advanced computer vision approach to detect recyclability is main task of the work and dataset to experiment on novel approaches are obtained from a student work from Stanford University. The dataset from Stanford clearly marks itself out due to its distinct features and the preprocessing that has been done. The dataset has been used by many of researchers in making more predictions and analysis as well. The size of the dataset being small needs to get some additional data in order to increase its accuracy as well as reliability.

\medskip

\section{Dataset Preparation}

A total of 1800 waste images are taken from the Stanford TrashNet Dataset[3]. The images are clicked on the white background and necessary pre-processing steps were done. We took 4 labels for classification i.e. Paper, Plastic, Metal, Glass as shown in Fig. 2. A total number of 1302 local waste images are captured using the mobile phones for four classes of data. A total of 3102 images are used for analysis and classification of the waste. The data are approximately divided equally in four classes, which helps to reduce the skewness in the data. The total number of image data is represented in Table I.

\par
The next step after collecting the data is the pre-processing step needed for cleaning the data. Various pre-processing steps can be utilized in order to clean data and make it ready to feed into the network. The real world data is random and has much noise. The feeding of image data into the network without performing the initiative step of data pre-processing would reduce the performance of the network.

 \begin{figure}[hbt!]
  \centering
  \includegraphics[scale=0.40]{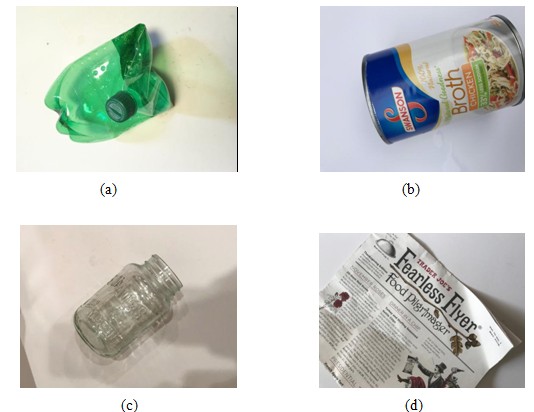}
  \caption{Dataset(a) Plastic (b) Metal (c) Glass (d) Paper [3]}
  \end{figure}

\section{Proposed System}

\subsection{Transfer Learning}
Deeper neural networks are more difficult to train due to necessity of high computation power. Due to the reason, we have used pre-trained model which was trained on ImageNet Dataset. As the number of images in our dataset are less and we have limited computational resources, transfer learning has been used in our system. Using transfer learning, the model seems to perform extremely well on our validation set after fine tuning the model. We have used various pre-trained model like ResNet[7] and VGG[8] to perform comparative analysis on those model. The losses and accuracy for each model are analysed properly. We have used ResNet18 architecture due to its better performance on our validation set.

\begin{table}
\centering
\caption{Number of images for each labels}
\begin{tabular}{|c|c|} 
\hline
\textbf{Labels} & \textbf{Total Images} \\ 
\hline
Plastic & 868 \\
Paper & 868 \\
Metal & 590 \\
Glass & 776 \\
\hline
\end{tabular}
\end{table}

 \begin{figure}[hbt!]
  \centering
  \includegraphics[scale=0.35]{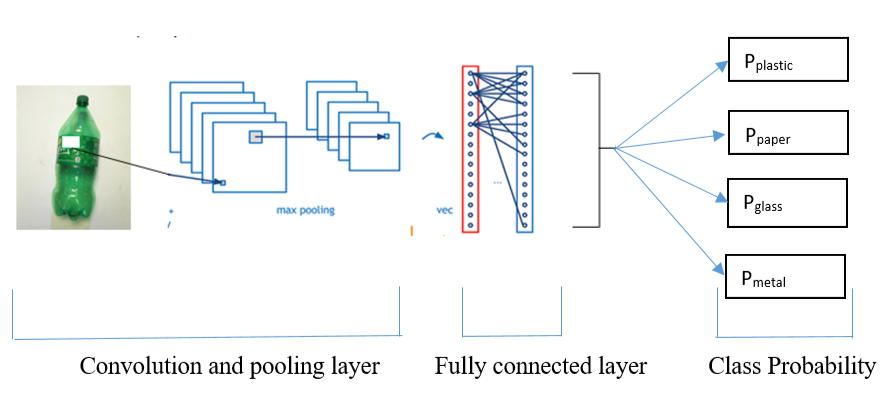}
  \caption{CNN Architecture}
  \end{figure}

\subsection{Image Preprocessing}

One of the first steps is to ensure that the images have the same size and aspect ratio. Most of the neural network models assume a square shape input image, which means that each image needs to be check if it is as square or not, and resized appropriately. Initially the size of image was extremely large. The size of image ranged from (2000-4000) pixels which cannot be fed into the convolutional layers of deep CNN model. If large number of input dimension of image is taken, then the training time would take longer and there would be unnecessary parameters or features to be learned for image classification. Hence, the image was resized to 512 pixels from (2000-4000) pixel for the training process as shown in Fig. 4.

 \begin{figure}[hbt!]
  \centering
  \includegraphics[scale=0.35]{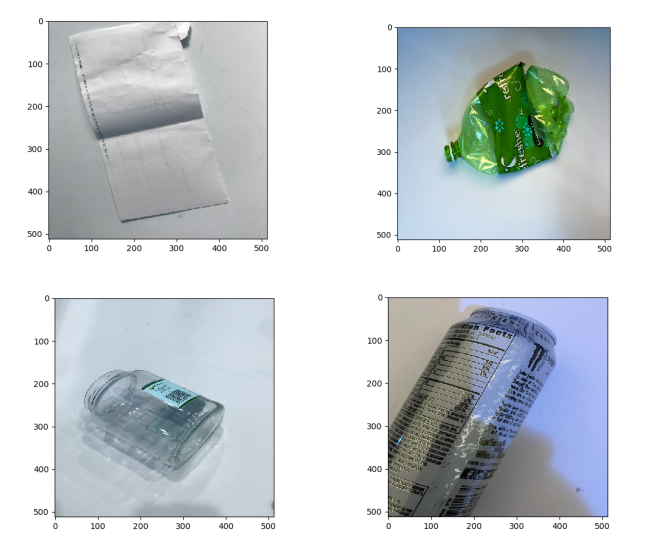}
  \caption{Resizing image into 512 X 512 pixels}
  \end{figure}

The mean values for each pixels across all the training examples was visualized which gave underlying structure in the images and an understanding about the whole dataset. Data normalization[9] is an important step, which ensures that each input parameter (pixel for images) has a similar data distribution. Normalization thus makes convergence faster while training the network. Data normalization is done by subtracting the mean from each pixel and then dividing the result by the standard deviation. The distribution of such data resemble a Gaussian curve centered at zero. For image inputs, individual pixel value needs to be positive, so the scale of [0, 255] was used to scale the normalized data as shown in Table II.

\begin{table}
\centering
\caption{Normalized mean and standard deviation for each channels}
\begin{tabular}{|c|c|c|} 
\hline
\textbf{Channels} & \textbf{Normalized Mean} & \textbf{Normalized Standard Deviation} \\ 
\hline
Red & 0.6067545935423009 & 0.1829832537916729 \\
Green & 0.5852166367838588 & 0.1823482159344223 \\
Blue & 0.565204377558798 & 0.1926306225491462 \\
\hline
\end{tabular}
\end{table}

 A number of 3102 images are not sufficient in order to train the neural network and get improved performance. We have done different types of image augmentation[10] like horizontal flip, random crop, zoom etc. to make variations inside the data so that it can generalize the unseen data accurately.  The images after augmentation are shown in Fig. 5.

 \begin{figure}[hbt!]
  \centering
  \includegraphics[scale=0.35]{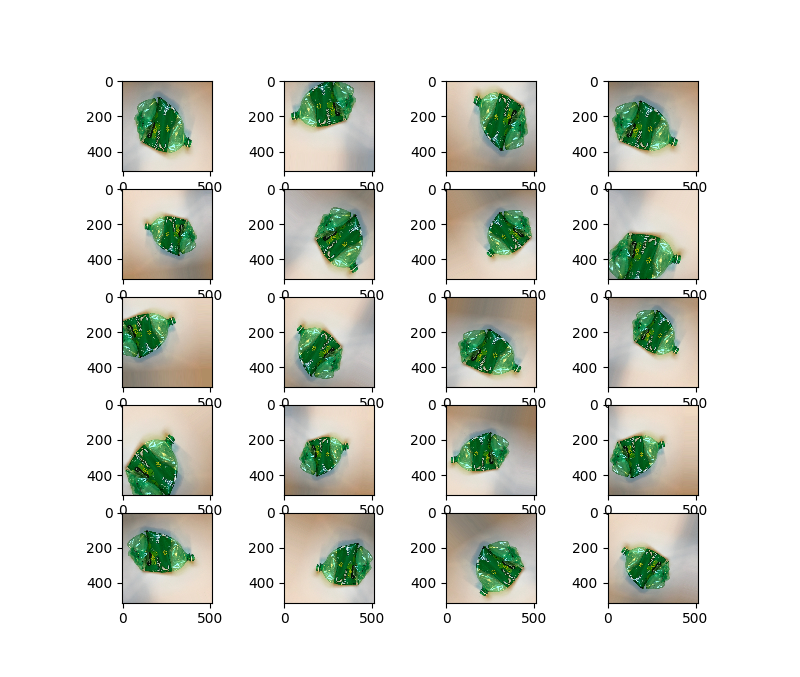}
  \caption{Data augmentation}
  \end{figure}

\section{Experiments}
We have implemented our method using Pytorch framework. Training was done on an NVIDIA GTX 1050 with 768 CUDA cores. The training took approximately three hours.  We trained across various pre-trained ImageNet Models and compare the metrics across each model. The feature extraction across images of various lightening, occlusion, illumination and resolution was performed. Then, after fine-tuning the model[11], we get the best performance metrics using ResNet18.

\subsection{Feature Extraction}
Feature extraction process is performed to extract interesting features from new samples learned by a previous network. So, we used convolutional layer of pre-trained network, ran the new data through it, and then trained a new classifier on top of the model. In the feature extraction process, we extract features by unfreezing the last three convolutional layers of pre-trained network then added our  fully connected layers and trained on our dataset. 

\subsection{Fine Tuning}
After performing the feature extraction, another model reuse technique which is identical to feature extraction is used called fine tuning. All the layers except last layers are frozen and a custom layer was built in order to perform the classification. In the last layer, two linear layer were added with ReLU activation function and finally the Softmax activation function. Then the convolutional layer and newly added classifier are jointly trained which improved model performance after fine tuning.

\subsection{Training}
The dataset has been divided into two part before building the model. 80\% of the total images are considered as the training images and remaining 20\% images is considered as validating images. The training images are used to train the network. The validation image is a sample of data held back from training the model that is used to give an estimate of model skill while tuning model’s hypothesis. The evaluation of a model skill on the training dataset would result in a biased score. So, the model is evaluated on the held-out sample to give and unbiased estimate of model skill. So, on loading the dataset, 20\% of the dataset is chosen as validation set to evaluate the model. The validation set is not used to train the data and only used to check model performance. Negative Log Likelihood loss function is used for determining the loss criteria. In the same way, Adam optimizer was used for optimizing the parameters with the learning rate of 0.001 to get the optimized minimum value using gradient descent technique. After training the model, accuracy was noted. In addition to model’s accuracy, the training loss and validation loss were also noted.

\section{Results and Analysis}

\subsection{Extraction of Feature Map}
The feature map extraction part will be extremely important for well predicting of the images. Model building exhibits many layers, which extracts the features of the image while training.The features are extracted when the filters are convolved with the images after passing through each layers. Edges, Lines, parts of objects are extracted from the image during initial stages of neural network for the actual classification as shown in Fig. 6.

\begin{figure}[hbt!]
 \centering
  \includegraphics[scale=0.4]{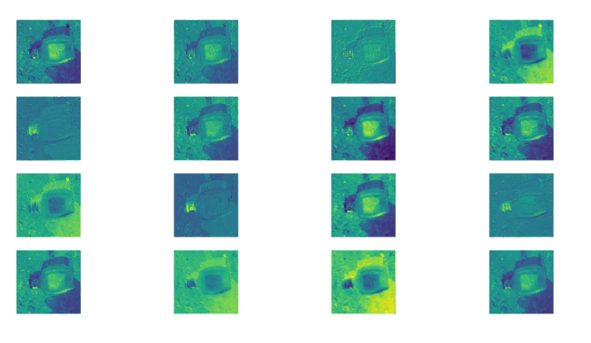}
  \caption{ Feature extraction on initial layers of CNN   }
\end{figure}

The filters used in the next layers extract more prominent features of the images after some extraction of primary features in initial layers of CNN, which help to distinguish the actual images properly as shown in Fig. 7.

\begin{figure}[hbt!]
 \centering
  \includegraphics[scale=0.35]{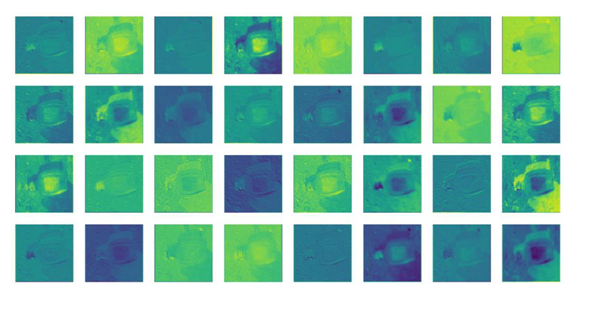}
  \caption{   Feature extraction using convolution after initial layers   }
\end{figure}

The last layers of CNN also extract the features like deformable shape analysis for the prediction of image as shown in Fig. 8. 

\begin{figure}[hbt!]
 \centering
  \includegraphics[scale=0.35]{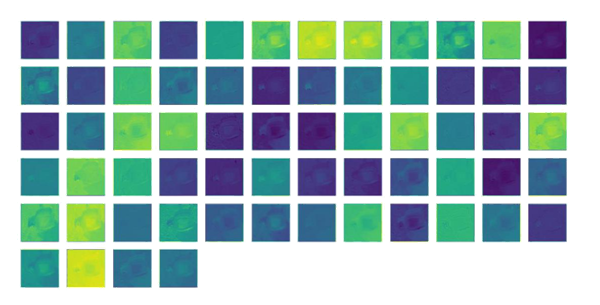}
  \caption{  Feature extraction across the last layers of the CNN   }
\end{figure}

 \subsection{Comparative Analysis of Different Models }
 Comparison between several model’s strengths and weaknesses were clearly observed before finalizing which model should be used as to applicable standards. The final chosen model was ResNet18 to be deployed into the hardware to make some real time predictions in the system.

\begin{enumerate}

\medskip
\item Performance on ResNet50
 
Fig. 9. shows a consistent difference between the training and validation loss curve in the model. The training error was decreased consistently but the model’s generalization capacity was not very good. In short, the model can be understood as being too much dependent on the dataset for prediction than generalization.

\begin{figure}[hbt!]
 \centering
  \includegraphics[scale=0.45]{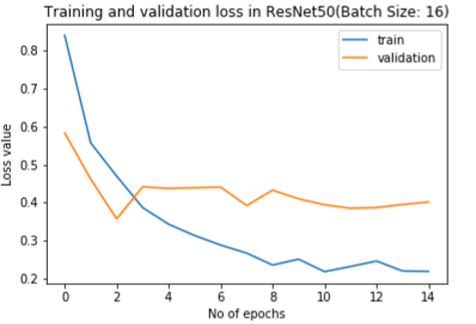}
  \caption{  Training and Validation Loss for ResNet50  }
\end{figure}

\item Performance on VGG16

VGG16 model was observed characteristically similar to the ResNet50 as it also had capacity of representing more complex model than the prepared dataset required. The loss curve was observed as overfitting since the training loss was observed to be around 0.2 and still decreasing as shown in Fig. 10. However, the model’s validation did not improve with reduction in training loss.

\begin{figure}[hbt!]
 \centering
  \includegraphics[scale=0.45]{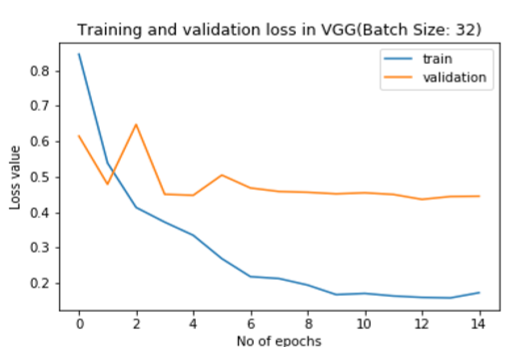}
  \caption{  Training and Validation Loss for VGG16  }

\end{figure}

\item Performance on ResNet18

The ResNet18 model seem to be good after visualizing the accuracy of both training and validation. The condition of overfitting and under fitting did not occur significantly in the model. he training loss as well as validation loss was decreasing exponentially which conclude that the network is good for predicting the test image and perform the classification task. Fig. 11. shows the number of epochs versus Loss value. The total number of epochs was taken 25 and the training and validation loss started to decrease initially. After some epoch, the loss was constant. It means that the model stop to learn from the corresponding epoch and start to saturate. The graph depicted that it was not necessary to run the model to train for many epochs more than five since the growth of the validation accuracy stopped after epoch 6. The validation accuracy seemed reasonable which showed that the model was neither overfitted nor underfitted as shown in Fig. 12.

\begin{figure}[hbt!]
 \centering
  \includegraphics[scale=0.45]{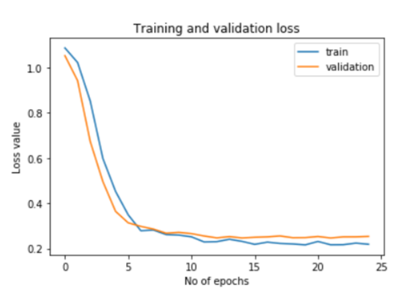}
  \caption{  Training and Validation Loss for ResNet18 }
\end{figure}

\begin{figure}[hbt!]
 \centering
  \includegraphics[scale=0.45]{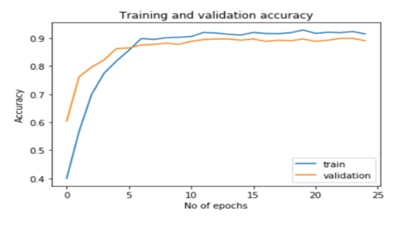}
  \caption{  Training and Validation Accuracy for ResNet18  }
\end{figure}

\end{enumerate}

 \subsection{Confusion Matrix }
 
The confusion matrix in Table III. shows that paper are mostly classified correctly more than any other type of waste. In the confusion matrix, index of each row corresponds to a true label and the indices of each column denotes the predicted label. The color of each cell represents the probability of the prediction and the number appearing on each cell gives the occurrences of the prediction. The predicted label is equal to the true label which is represented by leading diagonal of the confusion matrix, while off-diagonal elements are those that are mislabeled by the classifier.The higher diagonal values of the confusion matrix indicate more correct predictions. The glass can be seen as being predicted as plastic frequently. The situation can be explained because the transparency present in glass and plastic can resemble them to be detected in each other’s class now and then. Regarding very low correct prediction rate of plastic is due to its variation in size, shape and color frequently. Most of the things that are visible around us are made from plastic. The metal has significantly higher degree of correct prediction due to the size of dataset. The dataset has incorporated metal scraps, metal cans and so on thereby representing most of the metal products that are seen around. The paper is also significantly seen as plastic because of the fact that they are similar, colored or not, sheet like appearance and often deceive humans as well.  Through analysis via prediction matrix, the model has been found to have significant comparisons to human intervention and working. The features used by the model seem to be similar to human comparison between wastes since the confusion matrix shows confusion in those places that confuse humans as well. 
 
\begin{table}
\centering
\caption{Confusion matrix for ResNet18 architecture}
\begin{tabular}{|c|cccc|} 
\hline
 & \textbf{Glass} & \textbf{Metal} & \textbf{Paper} & \textbf{Plastic} \\ 
\hline
\textbf{Glass} & \textbf{83.8} & 3.7 & 3.7 & 8.8 \\
\textbf{Metal} & 1.7 & \textbf{89.0} & 2.5 & 6.8 \\
\textbf{Paper} & 0.1 & 1.1 & \textbf{90.2} & 8.6 \\
\textbf{Plastic} & 7.5 & 4.6 & 5.2 & \textbf{82.8} \\
\hline
\end{tabular}
\end{table}

The results show that the model showed mostly correct predictions. Fig. 13. showed correctly classified and misclassified images on test data. The misclassification is mainly because of the underlying disturbance in the picture. The pictures down sampling had reduced the clarity for pictures. In addition, due to inherent similarity between paper and plastic, the model could not correctly predict such case.

\begin{figure}[hbt!]
 \centering
  \includegraphics[scale=0.45]{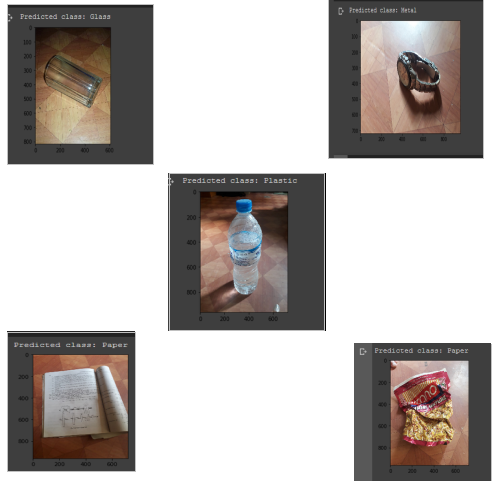}
  \caption{ Some of the correctly classified and misclassified images  }
\end{figure}

\section{Conclusion}

Through this research, we obtained the accuracy above 87\% and  made the comparative analysis on the model we adapted to train our data. For this, we have been putting a continuous effort for getting better result for each evaluation metrics. Furthermore, we still have things to work on dataset and increase its number. Also, the performance of model can be improved further by increasing the number of images and fine tuning the model properly. The idea of waste classification can be used in recycling process of waste if the mode is reliable and more accuracte.

\medskip
\section*{{Acknowledgment}}
 We would like to show gratitude and heartily thank our supervisor Dr. Surendra Shrestha, Head of Department of Electronics and Computer Engineering for his support, motivation and guidance for enhancing our idea into a useful product. His guidance has been the key to what we have achieved today. We would also like to thank to the Lab Assistant Mr. Kamal Nepal for his help in making decisions about enhancements on the hardware side and the design of our hardware module.
We would also like to thank the Department of Electronics and Computer Engineering IOE for giving us an amazing opportunity to carry out major project that has helped us to implement theoretical knowledge we gain from our study into the practical field and implement it in real life scenario.


\end{document}